\documentclass[10pt, a4paper]{article}
\usepackage{lrec2022} 
\usepackage{multibib}
\newcites{languageresource}{Language Resources}
\usepackage{graphicx}
\usepackage{tabularx}
\usepackage{todonotes}
\usepackage{comment}
\usepackage{epstopdf}
\usepackage[utf8]{inputenc}

\usepackage[hyphens]{url}
\usepackage{hyperref}
\usepackage{xstring}

\usepackage{color}
\usepackage{rotating}

\usepackage{xspace}
\newcommand{\Gerbil}{Gerbil}

\newcommand{\wrt}{w.r.t.~}
\newcommand{\st}{s.t.,~}
\newcommand{\eg}{e.g.,~}

\newcommand{\cf}{cf.,~}

\newcommand{\QAKiS}{QAKiS\xspace}
\newcommand{\TeBaQA}{TeBaQA\xspace}
\newcommand{\Elon}{Elon\xspace}
\newcommand{\QASystem}{QASystem\xspace}
\newcommand{\gAnswer}{gAnswer\xspace}

\newcommand{\QAmp}{QAmp\xspace}

\newcommand{\URI}{URI\xspace}

\newcommand{\Fscore}{F1 Score\xspace}
\newcommand{\KGQA}{KGQA\xspace}
\newcommand{\QA}{QA\xspace}
\newcommand{\QALD}{QALD\xspace}
\newcommand{\LCQuAD}{LC-QuAD\xspace}

\newcommand{\Wikidata}{Wikidata\xspace}

\newcommand{\DBpedia}{DBpedia\xspace}

\newcommand{\SPARQL}{SPARQL\xspace}

\usepackage{titlesec}
\titleformat{\section}{\normalfont\large\bfseries\center}{\thesection.}{1em}{}
\titleformat{\subsection}{\normalfont\SmallTitleFont\bfseries\raggedright}{\thesubsection.}{1em}{}
\titleformat{\subsubsection}{\normalfont\normalsize\bfseries\raggedright}{\thesubsubsection.}{1em}{}
\renewcommand\thesection{\arabic{section}}
\renewcommand\thesubsection{\thesection.\arabic{subsection}}
\renewcommand\thesubsubsection{\thesubsection.\arabic{subsubsection}}

\def\numberOfEvaluatedPapers{100~\xspace} 
\def\numberOfEvaluatedSystems{98\xspace} 
\def\numberOfEvaluatedDatasets{4\xspace}
\def\numberOfRepositories{24\xspace}
\def\numberOfDemosAndAPIs{16\xspace}
\def\numberOfDemosAndAPIsWorking{8\xspace}

\title{Knowledge Graph Question Answering Leaderboard: \\A Community Resource to Prevent a Replication Crisis}

\name{A. Perevalov, Xi Yan, L. Kovriguina, L. Jiang, A. Both and R. Usbeck} 

\address{Anhalt University of Applied Sciences, Fraunhofer IAIS, University Hamburg \\
\{aleksandr.perevalov, andreas.both\}@hs-anhalt.de, \\
liubov.kovriguina@iais.fraunhofer.de,\\
\{xi.yan, longquan.jiang, ricardo.usbeck\}@uni-hamburg.de,}

\abstract{
Data-driven systems need to be evaluated to establish trust in the scientific approach and its applicability.
In particular, this is true for Knowledge Graph (KG) Question Answering (QA), where complex data structures are made accessible via natural-language interfaces. 
Evaluating the capabilities of these systems has been a driver for the community for more than ten years while establishing different KGQA benchmark datasets.
However, comparing different approaches is cumbersome. 
The lack of existing and curated leaderboards leads to a missing global view over the research field and could inject mistrust into the results. 
In particular, the latest and most-used datasets in the KGQA community, LC-QuAD and QALD, miss providing central and up-to-date points of trust. 
In this paper, we survey and analyze a wide range of evaluation results with significant coverage of \numberOfEvaluatedPapers publications and \numberOfEvaluatedSystems systems from the last decade. 
We provide a new central and open leaderboard for any KGQA benchmark dataset as a focal point for the community - \href{https://kgqa.github.io/leaderboard/}{https://kgqa.github.io/leaderboard/}.
Our analysis highlights existing problems during the evaluation of KGQA systems.
Thus, we will point to possible improvements for future evaluations.
\\ \newline }

\begin{document}

\maketitleabstract

\section{Introduction}
Question Answering (\QA) is a rapidly growing field in research and industry\footnote{\href{https://www.gartner.com/smarterwithgartner/2-megatrends-dominate-the-gartner-hype-cycle-for-artificial-intelligence-2020}{https://www.gartner.com/smarterwithgartner/2-megatrends-dominate-the-gartner-hype-cycle-for-artificial-intelligence-2020} (September 28, 2020)}.
\QA systems already deliver their potential into many real-world problems, \eg~\cite{app11125456,DBLP:conf/i-semantics/BothPBHIHSWFG21,diefenbach2021wikibase}.
These systems can be divided into two main paradigms~\cite{jurafsky2018speech}: IR-based that works over unstructured data, closely related to Machine Reading Comprehension and Retriever-Reader architecture) and Knowledge-Based (KBQA) which works over structured data, such as relational tables, specific data APIs, knowledge graphs (KGs).
In this regard, Question Answering over Knowledge Graphs (KGQA) is of particular interest to this work.

Many different benchmarking datasets are evaluating KGQA systems.
These datasets differ in the underlying knowledge graph (\eg \DBpedia~\cite{auer2007dbpedia} or \Wikidata~\cite{wikidata}), size order of magnitude~\cite{DBLP:journals/corr/abs-2007-13069}, questions complexity~\cite{saleem2017question}, multilingual support~\cite{chandra2021survey}, and many more dimensions.
In the KGQA research community, several datasets have become a de facto standard for evaluation of such systems, such as the QALD~\cite{DBLP:conf/semweb/UsbeckGN018} and \LCQuAD~\cite{DBLP:conf/semweb/DubeyBA019} benchmark dataset series.
As more and more researchers introduce new evaluation results using these well-known datasets, it becomes more challenging to follow the up-to-date state-of-the-art in the KGQA field.
The related research fields such as IR-based QA and Knowledge Graph research community already have their own well-established and maintained leaderboards of the best solutions (SQuAD\footnote{\cf The Stanford Question Answering Dataset
leaderboard at \href{https://rajpurkar.github.io/SQuAD-explorer/}{https://rajpurkar.github.io/SQuAD-explorer/}}~\cite{DBLP:conf/emnlp/RajpurkarZLL16}, OGB\footnote{Open Graph Benchmark -- is a collection of the benchmark datasets for machine learning over graphs: \href{https://ogb.stanford.edu}{{https://ogb.stanford.edu}}} \cite{hu2020open}). However, it is not the case for KGQA.
This lack - in particular of curated leaderboards - leads to a missing global view over the research field.
In turn, this could inject mistrust into result tables within publications when they are incomplete or lack a comparison to certain systems, as often required by reviewers.
In particular, the latest and most-used datasets in the Semantic Web community, \LCQuAD and QALD, miss providing central points of trust such as leaderboards. 
In this paper, we analyze the publications of \KGQA evaluations of the last decade. 
We evaluated \numberOfEvaluatedPapers papers and \numberOfEvaluatedSystems systems on \numberOfEvaluatedDatasets datasets focusing on the LC-QuAD and QALD series.
Our results show that evaluation numbers are often consistent. 
Existing errors stem from minor differences in the data (\eg gAnswer~\cite{gAnswer} on QALD-9~\cite{DBLP:conf/semweb/UsbeckGN018}) that seems to be rounding errors or inconclusive behavior.
Finally, we discuss the consequences of our findings and will point to possible improvements for future evaluations.

Our contributions are as follows:
\begin{itemize}
\item We present the first, extensive evaluation analysis of the state of the research in \KGQA. 
\item We provide a new central and open leaderboard for any KGQA benchmark dataset as a focal point for the community - \href{https://github.com/KGQA/leaderboard}{https://github.com/KGQA/leaderboard}.
\item We provide an up-to-date overview of all available demos or Web services for KGQA at the point of publication.
\end{itemize}
These contributions should help the scientific community to foster replication and cross-evaluation in the future.

In the following, we analyze related studies and approaches in Section 2. 
Afterward, we introduce the analyzed datasets and systems in Section 3. 
In Section 4, we describe our extensive state-of-the-art data and delve into its analysis. 
Next, we discuss possible interpretations and paths forward and end with a summary and outlook in Section 6. 
\section{Related Work}

There are multiple approaches to tracking the progress of any research field. In machine learning and NLP, these approaches can be subdivided into \textit{benchmarking frameworks} and \textit{manual or semi-automatic reporting platforms}. 

Today, benchmarking frameworks need to limit their scope to a subset of tasks to cover the necessary metrics and experiments types out-of-the-box. A general benchmarking framework, which works without writing code, does not exist. 
For KGQA, different \textit{benchmarking} frameworks have been proposed. For example, GERBIL QA~\citelanguageresource{DBLP:journals/semweb/UsbeckRHCHNDU19}, can benchmark KGQA systems via their Web APIs in a FAIR way~\cite{wilkinson2016fair}. It also has an integrated leaderboard\footnote{\href{http://gerbil-qa.aksw.org/gerbil/overview}{http://gerbil-qa.aksw.org/gerbil/overview}} which displays a summary of all experiments run via the platform. At the same time, this is the biggest downside - only experiments run via the platform are integrated. Thus, a realistic view depends on the adoption of the platform. This adoption seems to lack due to missing developer resources, which continuously update available systems and datasets. 
A different direction is followed by systems like \href{https://github.com/AKSW/irbench}{https://github.com/AKSW/irbench} or QALDGen~\cite{DBLP:conf/semweb/Singh0NCPN019}, which provide command-line tools for benchmarking any KGQA system. However, the offline nature of these tools leads to offline results, i.e., the results might be used in papers but do not contribute to a trustworthy overview of the field of research. 

Recently, \textit{reporting platforms} gained popularity. 
They allow quick access to results, but either they are curated manually via a community or semi-automatically updated.
\href{https://nlpprogress.com/}{https://nlpprogress.com/} is a famous community website launched by Sebastian Ruder. Regarding KGQA, the website's most recent information is 3 years old, possibly displaying the disinterest of the NLP community in semantic tasks.
\href{https://paperswithcode.com/}{https://paperswithcode.com/} is another reporting platform run by Facebook AI research allowing to openly edit papers, code, datasets, methods, and evaluation tables. While this is of tremendous help for reproducibility, its results for KGQA are sparse. There is only one result for LC-QuAD 2 and QALD 9, and both are for relation extraction rather than Question Answering. 

A promising semi-automatic approach is the Open Research Knowledge Graph (ORKG)~\cite{AuerOelenHarisStockerDSouzaFarfarVogtPrinzWiensJaradeh+2020+516+529} . It allows the community to persistently annotate papers via smart tools with meta and evaluation data, e.g., \href{https://www.orkg.org/orkg/paper/R6386/R6393}{https://www.orkg.org/orkg/paper/R6386/R6393} for QALD-6 data. However, the current adoption in the community does not go beyond prototypes provided by the ORKG team. 
A change might come with the  European Open Science Cloud (EOSC) and the Nationale Forschungsdateninfrastruktur für Data Science und Künstliche Intelligence (German National Data Infrastructure for Data Science and AI). Those publicly funded initiatives strive to foster ecosystems like ORKG in the long term. 

Finally, surveys can be viewed as reporting platforms. Different surveys have been published in the past decade focusing on a variety of topics such as challenges in general KGQA~\cite{DBLP:journals/semweb/HoffnerWMULN17}, challenges in complex KGQA~\cite{DBLP:journals/corr/abs-2007-13069}, core techniques of KGQA~\cite{DBLP:journals/kais/DiefenbachLSM18} or neural network-based KGQA systems~\cite{DBLP:journals/widm/ChakrabortyLMTL21}. However, these are automatically outdated when published or focus only on a narrow subtopic.

Thus, there is the need for a central, dense, and open reporting platform focusing on KGQA, which provides trustworthy insights.

\section{Benchmark Datasets and Systems}
We surveyed 14 DBpedia-based KGQA benchmark datasets that were published in the last decade (\cf Section \ref{sec:QAbenchmarks}). In this paper, we consider \numberOfEvaluatedDatasets \KGQA datasets for an in-depth analysis. 
Requirements for selecting a dataset include usage for the evaluation of different systems, availability in English, relying on DBpedia (primarily) or Wikidata (knowledge bases, which are still maintained), and cited above 5 times. Our goal was to make sure that the chosen QA datasets are: up-to-date, close to a real-world setting, can be manually evaluated, and are vastly studied.
Note, we use benchmark datasets and dataset synonymous.

We took \numberOfEvaluatedSystems \QA systems into the consideration. They are collected manually from articles that include evaluation results on the considered benchmark datasets. The article search was conducted in two ways. First, we retrieved articles using a keyword search on Google Scholar. Specifically, the selection criteria were: published after 2019, and that titles satisfy:  ['question answering' AND ('semantic web' OR 'data web' OR 'web of data')]. The second method is to extract all articles which cite the benchmark dataset from Google Scholar either as direct citation or as URL to the location of the dataset. We removed duplicates and manually extracted the QA systems evaluated or referred to in the articles. This resulted in \numberOfEvaluatedPapers analyzed papers. Note, some systems are evaluated on a subset of the dataset or a dataset where the benchmark dataset is just a subset. We indicated such a difference in the leaderboard accordingly.

\subsection{\KGQA Datasets}\label{sec:QAbenchmarks}

The first dataset is \QALD which is multilingual dataset challenge series.
In QALD-8, there were 219 training question-answer pairs and 42 test data points respectively. It was the first edition to use GERBIL QA as a benchmarking platform~\cite{DBLP:journals/semweb/UsbeckRHCHNDU19}.
The newest instance -- \QALD-9~\cite{DBLP:conf/semweb/UsbeckGN018} -- contains 558 questions incorporating information of the DBpedia knowledge base\footnote{\href{https://www.dbpedia.org/}{https://www.dbpedia.org/}} where for each question the following is given: a textual representation in multiple languages, the corresponding \SPARQL query (over DBpedia), the answer entity \URI, and the answer type. 
The QALD series has a growing number of questions per edition and thus grows continuously in its expressiveness.
The dataset has become a staple for many research studies in \QA (\eg~\cite{DBLP:journals/semweb/HoffnerWMULN17,DBLP:journals/kais/DiefenbachLSM18}).

The second and third dataset is \LCQuAD.
\LCQuAD (version 1)~\cite{DBLP:conf/semweb/TrivediMDL17} is a Question Answering dataset with 5000 pairs of questions and its corresponding SPARQL query. \LCQuAD v2~\cite{DBLP:conf/semweb/DubeyBA019} is the follow-up dataset with 30.000 question-answer pairs to better suit novel machine learning approaches.
The SPARQL queries are intended to be executed on DBpedia. 
\LCQuAD is widely used in the process of QA systems development~\cite{DBLP:conf/www/SinghRBSLUVKP0V18,DBLP:conf/semweb/DubeyBCL18}.


Other KGQA datasets are Free917 \citelanguageresource{free917}, WebQuestions \citelanguageresource{webquestions}, ComplexQuestions \citelanguageresource{complexquestions}, SimplesQuestions \citelanguageresource{simplequestionfreebase}, GraphQuestions \citelanguageresource{graphquestions}, WebQuestionsSP \citelanguageresource{webquestionssp}, 30MFactoidQA \citelanguageresource{30mfactoidqa}, ComplexWebQuestions \citelanguageresource{complexwebquestions}, PathQuestion \citelanguageresource{pathquestion}, MetaQA \citelanguageresource{metaqa}, TempQuestions \citelanguageresource{tempquestions}, TimeQuestions \citelanguageresource{timequestions}, CronQuestions \citelanguageresource{cronquestions}, FreebaseQA \citelanguageresource{freebaseqa}, Compositional Freebase Questions (CFQ) \citelanguageresource{cfq}, Compositional Wikidata Questions (CWQ) \citelanguageresource{cwq}, RuBQ \citelanguageresource{rubq1,rubq2}, GrailQA \citelanguageresource{grailqa}, Event-QA \citelanguageresource{eventqa}, SimpleDBpediaQA \citelanguageresource{simpledbpediaqa}, CLC-QuAD \citelanguageresource{clcquad}, KQA Pro \citelanguageresource{kqapro}, SimpleQuestionsWikidata \citelanguageresource{simplequestionswikidata}, DBNQA \citelanguageresource{dbnqa}, etc. 

These datasets do not fulfill our current criteria and thus are not part of the initial version of the KGQA leaderboard. However, we encourage the community to help us update the leaderboard also for these datasets to prevent a replication crisis before it starts.
\subsection{\QA systems}\label{sec:QAsystems}
While there are decentral collections of KGQA systems and there are available as code or Web service, \eg \href{https://github.com/semantic-systems/NLIWOD/tree/master/qa.systems}{https://github.com/semantic-systems/NLIWOD/tree/master/qa.systems}, there is no up-to-date and systematically curated collection as of now.
Our analysis shows that \numberOfRepositories provide a URL to a repository and \numberOfDemosAndAPIs even to an online demo or Web API. 
However, after inspection, only \numberOfDemosAndAPIsWorking demos or Web APIs are still functional. 
This is the first hint towards an upcoming replication crisis.
For a full list of systems, their descriptions, and pointers to their web services and demo, see \url{https://github.com/KGQA/leaderboard/blob/gh-pages/systems.md#Systems}

\section{Dataset Analyses}
\begin{figure*}[htb!]
    \centering
    \includegraphics[width=\textwidth]{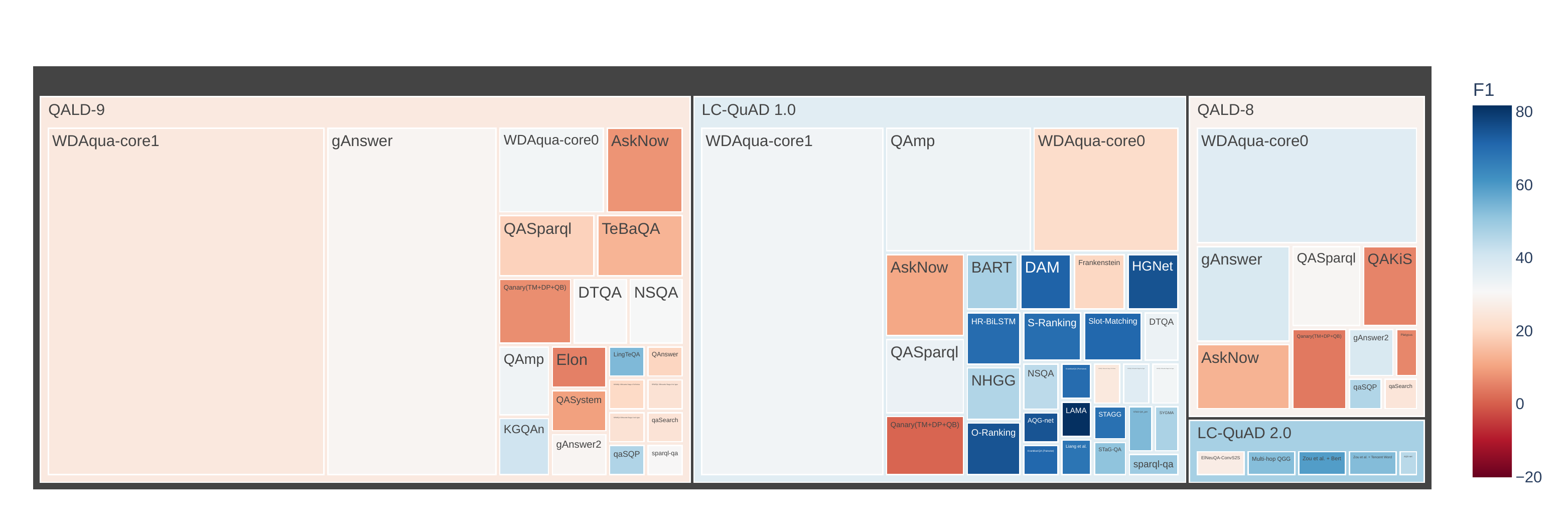}
    \caption{Treemap chart based on the collected results grouped by considered datasets (\QALD-8, \QALD-9, \LCQuAD 1.0, \LCQuAD 2.0). The \KGQA systems are located within the dataset rectangles. The size of the rectangles is proportional to the number of mentions of a particular system in the whole leaderboard. The color of the rectangles denotes the average \Fscore of the corresponding systems.}
    \label{fig:tree-map-chart}
\end{figure*}

We evaluated \numberOfEvaluatedPapers papers and \numberOfEvaluatedSystems systems focusing on \numberOfEvaluatedDatasets datasets, namely LC-QuAD version 1 and version 2 as well as the QALD-8 and 9 versions (all datasets released on 2017 or later).
Figure \ref{fig:tree-map-chart} comprehensively summarizes the considered results of the leaderboard.

\begin{figure*}[htb!]
    \centering
    \includegraphics[width=0.8\textwidth]{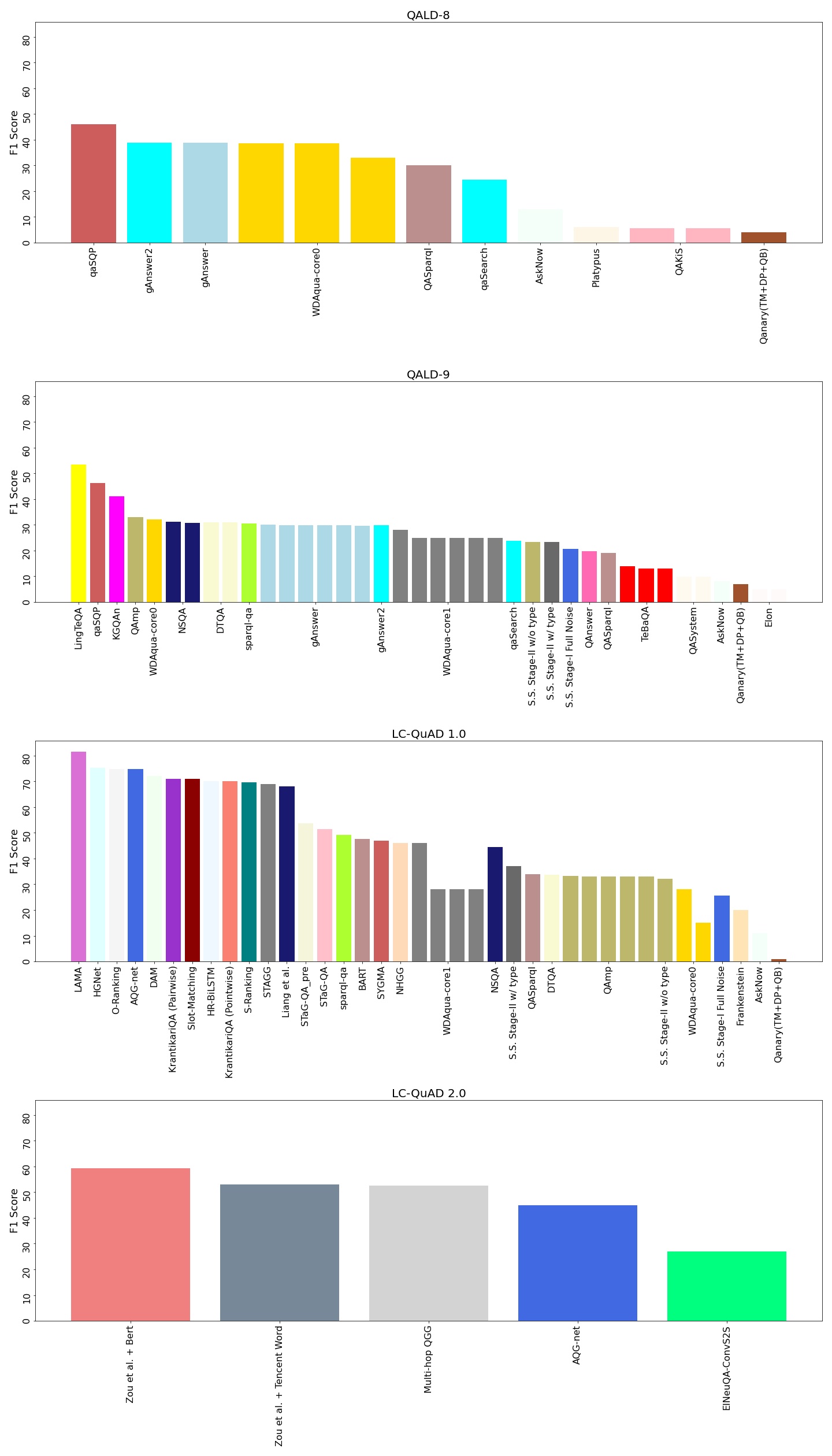}
    \caption{The chart demonstrates evaluation values (\Fscore) grouped by \KGQA systems (similar color) given a dataset. Each bar corresponds to a particular publication.}
    \label{fig:Grouped-bar-chart}
\end{figure*}

\begin{figure}[htb!]
    \centering
    \includegraphics[trim=0 50 98 0,clip,width=\linewidth]{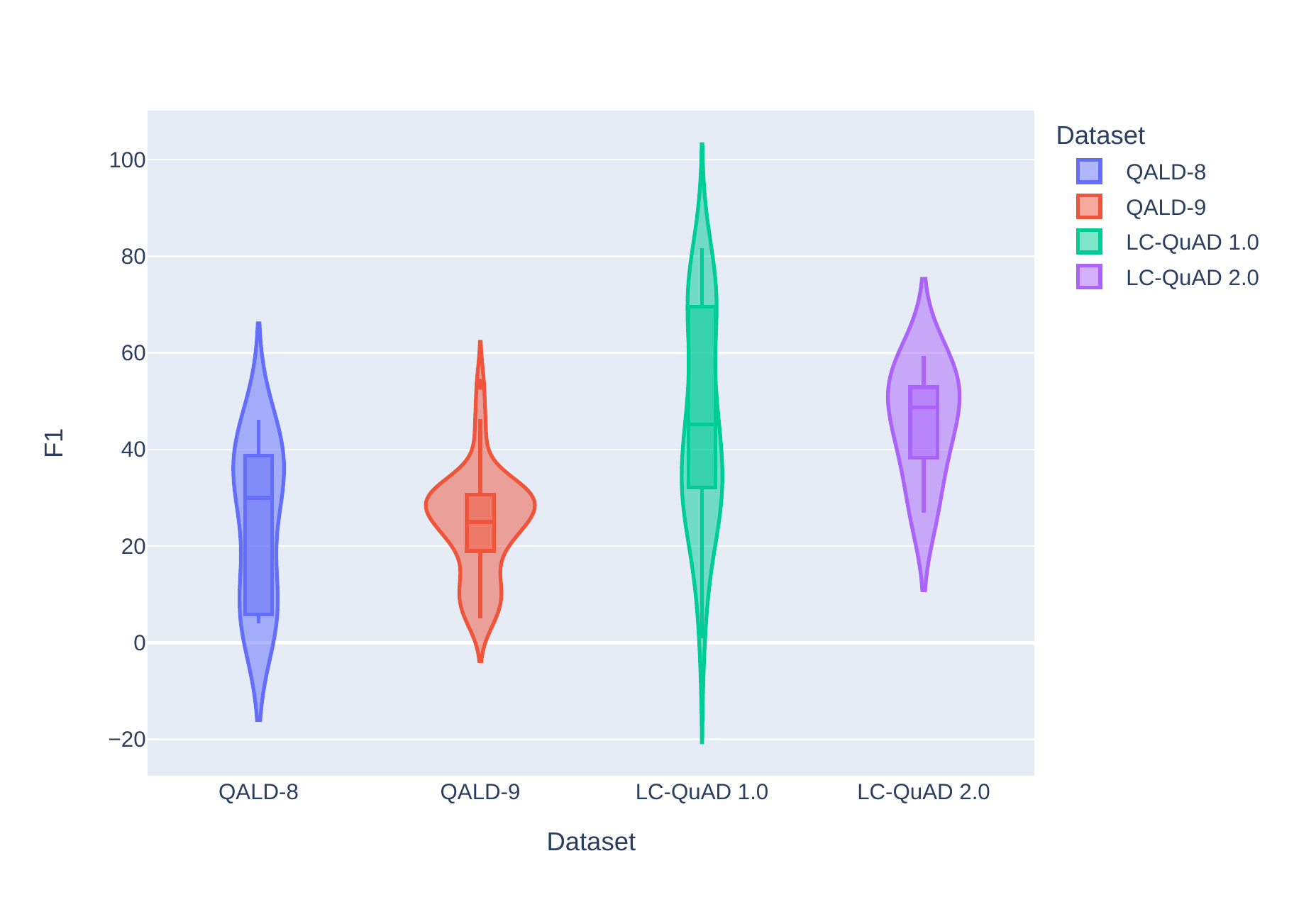}
    \caption{The figure demonstrates the distribution of the \Fscore values and their statistics from different publications given a dataset.}
    \label{fig:evaluation-of-datasets-violin}
\end{figure}

Based on the results, it became clear that \emph{the evaluation values across the publications are often consistent}.
The results contain multiple values for some of the system-dataset combinations (\eg WDAqua-core0 over \LCQuAD 1.0), reported by different publications.
Figure \ref{fig:Grouped-bar-chart} demonstrates the evaluation values given a particular benchmark dataset grouped by the \KGQA systems.
For system-dataset combinations with multiple values, we calculated the standard deviation (std.).
The std. values for such systems as \QAKiS, \TeBaQA, \Elon, \QASystem, \gAnswer, and \QAmp are not higher than 1\%.
This non-null std. is probably caused by the rounding errors.
The only outliers in the evaluation values were observed given the WDAqua-core systems.
For example, the paper \cite{Zheng2019QuestionAO} reports \Fscore of 38.7\% for WDAqua-core0 over \QALD-8, taking the results from the original publication of WDAqua-core0. 
Another paper \cite{Orogat2021CBenchTB} reports \Fscore of only 33.0\% for the same system-dataset combination. The authors \cite{Orogat2021CBenchTB} calculated this result.
The std. of both WDAqua-core versions reaches 9\% on \LCQuAD 1.0 dataset and 3\% on \QALD-8.
Note, the high std. values are not dependent on the datasets.
Hence, the papers reporting significantly different results regarding WDAqua-core require further investigation.
One of the assumptions is that WDAqua-core provides a publicly accessible demonstrator and API\footnote{\href{https://qanswer-frontend.univ-st-etienne.fr}{https://qanswer-frontend.univ-st-etienne.fr}} which enables researchers to re-run the evaluation.
This fact naturally implies possible differences in the evaluation results.
However, there is no such systematic tendency for the other results as probably the majority of them were not reproduced but cited from the original publication.
Despite the consistency of the results, the \Fscore values of the systems have a wide variance range given a particular dataset (\cf Figure \ref{fig:evaluation-of-datasets-violin}). 

Surprisingly, the number of papers from ArXiv (pre-prints) in our leaderboard appears to be higher than the number of peer-reviewed papers (54\% vs 46\%).
It was observed that the peer-reviewed papers report significantly higher results \wrt \Fscore which is 30.2\% for preprints and 39.5\% for peer-reviewed papers.
The logical reason for this is that the peer-reviewed papers typically report state-of-the-art results, while preprints might contain preliminary work.

Given the considered results, it was observed that the authors of 72\% papers did not include all the evaluation results from other publications in their comparison that were already available at a particular point in time.
To find out this number, the set of systems from a publication reporting the values on particular datasets was compared to the set of systems released a year ago or earlier.
For example, the publication \cite{Orogat2021CBenchTB} released in 2021 does not consider the results of the \QAmp system \cite{vakulenko2019message} that was published in 2019.





\section{Discussion}
The trustworthiness of scientific results strongly depends on the comparability and replicability of the same.
In the field of KGQA, one could assume that the existence of a large and rising number \QA datasets ensures comparability.
Indeed, our analysis shows that the reported evaluations are \textit{overwhelmingly coherent}. 
However, we observed several issues:
First, the main reason why most numbers are identical is that people refer to results given in an original paper and its evaluation section. 
We could not find evidence that researchers actively tried to replicate results. 
A reason could be that only, \numberOfDemosAndAPIs percent of the systems are available as source code (or Web service/demo).
However, even in the existence of an online demo, \eg \cite{Diefenbach2017,diefenbach2020towards}, the current state of the \KGQA system seems not to be re-evaluated.

Second, our analysis indicates that researchers might have overlooked (best case) or omitted (worst case) relevant results that speak against their claims. 
For example, in~\cite{wu2021modeling} there are similar earlier works~\cite{Maheshwari2019,Nhuan2020} which evaluated the same datasets and provided similar or even better results.
However, we are well aware that researchers struggle with establishing an up-to-date overview of current research due to the time-consuming nature of the process without a central overview of \KGQA systems.

Third, we see a strong need for improved evaluation methods. 
This demand can be covered by online evaluation methods, e.g., using platforms like \Gerbil~\citelanguageresource{DBLP:journals/semweb/UsbeckRHCHNDU19}). 
However, we also observed a decreasing amount of working online demos suggesting that a new form of a platform where models as such can be uploaded\footnote{For example,  \href{https://project-hobbit.eu/outcomes/hobbit-platform/}{https://project-hobbit.eu/outcomes/hobbit-platform/}.} could be a future direction.

Fourth, while developing new platforms and systems, we should also consider the rising critique on leaderboards regarding their utility for the NLP community at large~\cite{DBLP:journals/corr/abs-2009-13888}. 
Thus, we concur that evaluation protocols need to be published to foster transparency on leaderboards.

Finally, the lack of open-source implementations could be a starting point for a replication crisis. 
While there is no replication crisis in the field of KGQA as of now, the community needs to leverage novel initiatives such as the European Open Science Cloud\footnote{\href{https://eosc-portal.eu/}{https://eosc-portal.eu/}} or the National Research Data Infrastructure for Data Science and AI\footnote{\href{https://www.nfdi4datascience.de/}{https://www.nfdi4datascience.de/}}. 
Otherwise, models and source code might be lost or results will become incomparable in the long term.
\section{Summary and Future Work}
In this paper, we presented a novel community resource to track advances in the field of KGQA research. 
We foresee the need to maintain a \KGQA focused platform as long as approaches such as ORKG~\cite{AuerOelenHarisStockerDSouzaFarfarVogtPrinzWiensJaradeh+2020+516+529} are not widely used or developed far enough. 
Of course, we could have just added our findings to reporting platforms. 
However, we believe, that this publication provides a more valuable base for discussions and reaches a wider audience than a silent upload. 
Additionally, since the QALD-9 evaluation campaign has passed for more than 3 years now, we intend to establish a central leaderboard to keep people on the same page.

In the future, we are looking into automatic ways to synchronize various reporting platforms with the KGQA leaderboard.
We plan to extend the evaluation of \QA systems, \st replicable evaluations, and data collections are possible. 
Additionally, improved metrics (\eg~\cite{DBLP:journals/pvldb/OrogatE21,Siciliani2021MQALDET}) should be evaluated over models, source code, or via platforms to allow in-depth analyses of the capabilities of \QA systems.

We are aware of research on other KGQA datasets grounded in Wikidata, Freebase, WikiMovies, and EventKG and want to encourage the community to update the KGQA leaderboard with the corresponding numbers.


\section{Bibliographical References}\label{reference}
\bibliographystyle{lrec2022-bib}
\bibliography{lrec2022-example}

\section{Language Resource References}\label{lr:ref}
\bibliographystylelanguageresource{lrec2022-bib}
\bibliographylanguageresource{languageresource}

\appendix

\section{KGQA Leaderboard}

To ensure the replication of KGQA systems and the trustworthiness of their evaluation results, we provide a leaderboard. The leaderboard is available at \href{https://kgqa.github.io/leaderboard/}{https://kgqa.github.io/leaderboard/}. It can be used to compare the capabilities of these KGQA systems over the latest and commonly used KGQA benchmark datasets by tracking the progress. It includes the datasets, links, papers and SOTA results.

\begin{figure}[htb!]
    \centering
    \includegraphics[width=\linewidth]{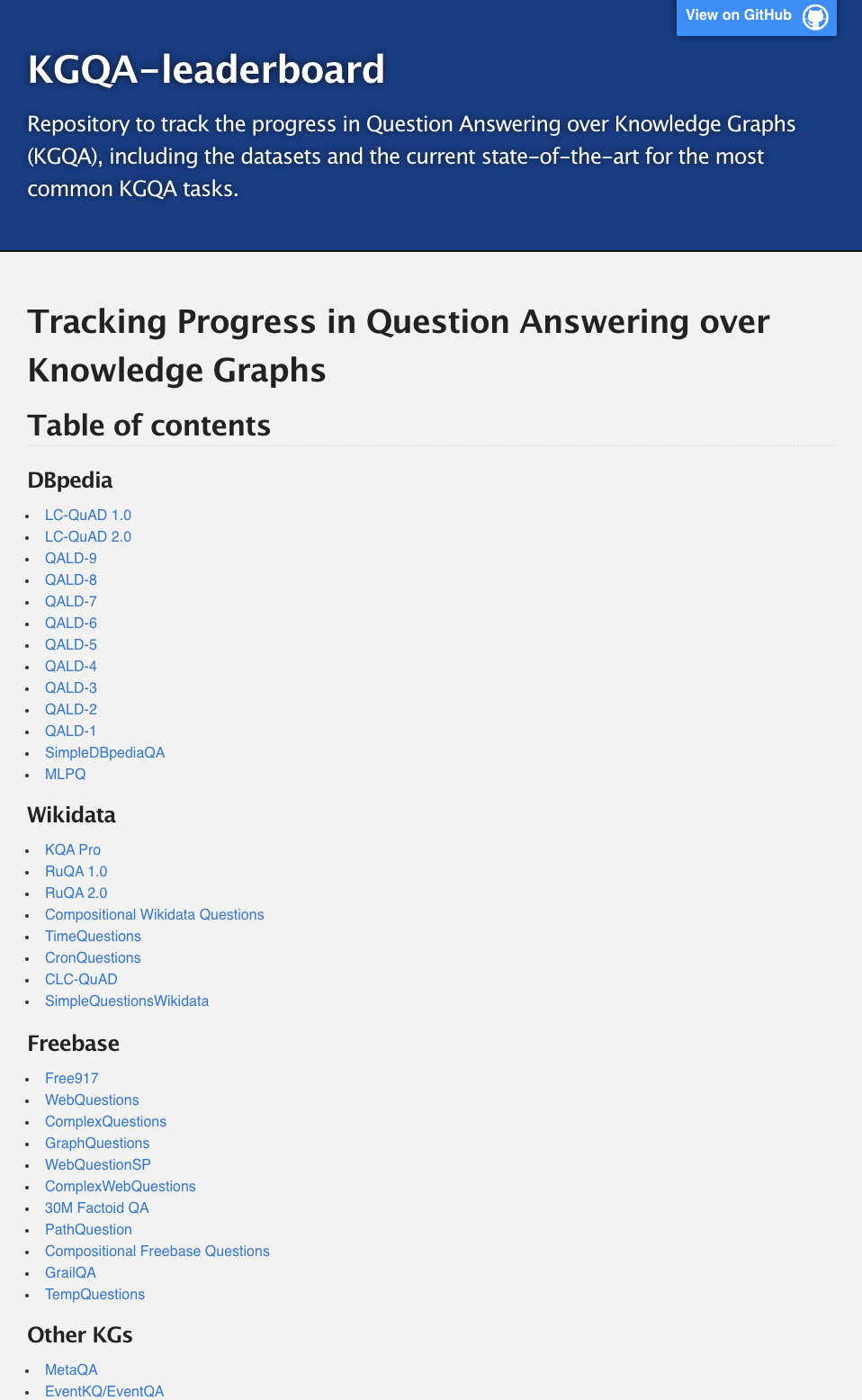}
    \caption{Interface of the KGQA leaderboard.}
    \label{fig:leaderboard}
\end{figure}

\begin{figure}[htb!]
    \centering
    \includegraphics[width=\linewidth]{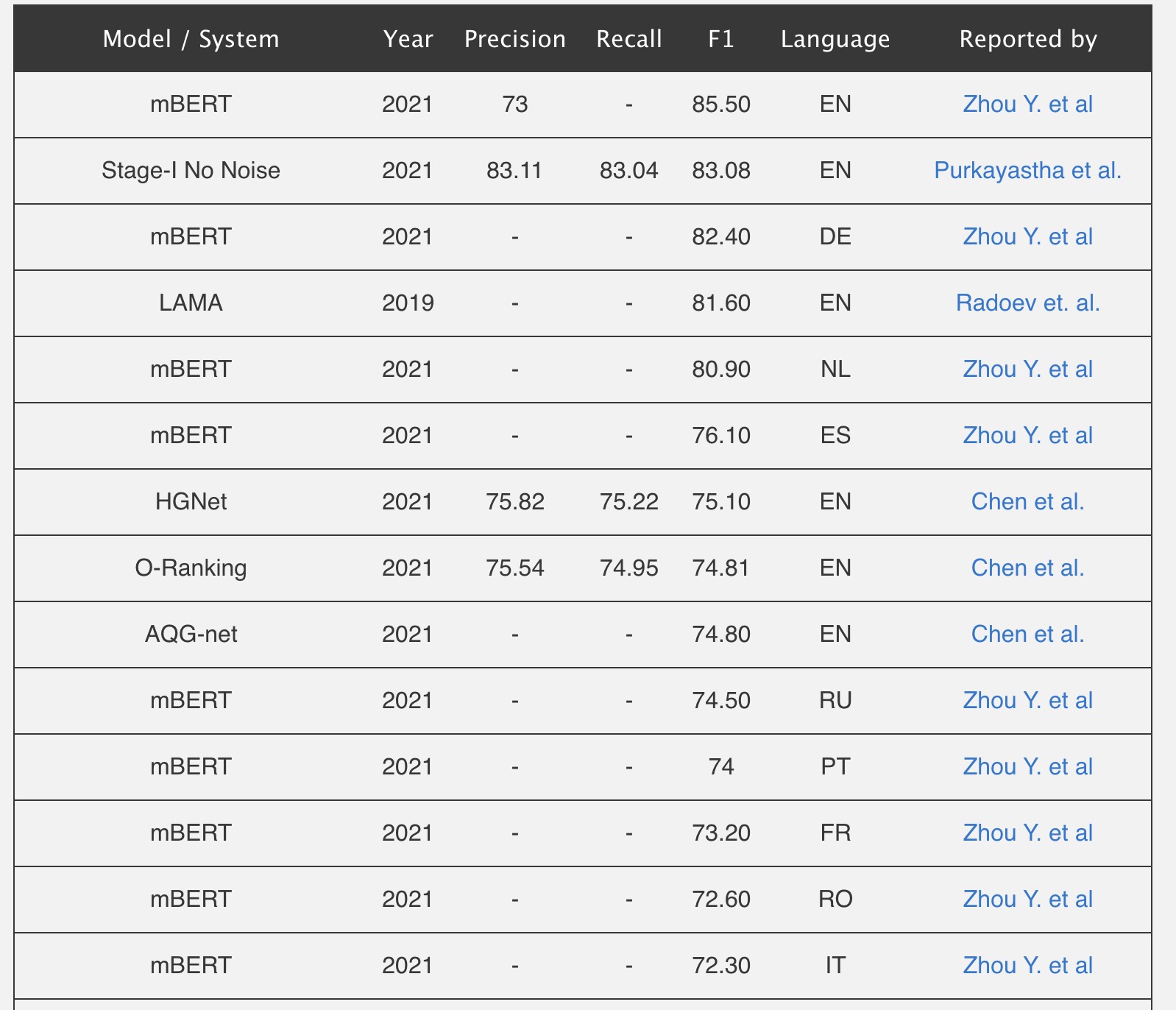}
    \caption{An example of \LCQuAD V1.0 Leaderboard.}
    \label{fig:lcquad-leaderboard}
\end{figure}

At the time of writing, the leaderboard includes a total of 34 KGQA datasets across 5 knowledge graphs (i.e., DBpedia, Wikidata, Freebase, WikiMovies, and EventKG). As shown in Fig.~\ref{fig:leaderboard}, these KGQA datasets are separated by the used target KGs. Fig.~\ref{fig:lcquad-leaderboard} shows an example of LCQuAD V1.0 Leaderboard. We will continuously add newly released datasets and their SOTA results, and invite other researchers to make their contributions by adding new results based on these KGQA dataset overviews.

\end{document}